\documentclass[conference,10pt,twocolumn]{IEEEtran}
\usepackage[dvips]{graphicx}
\usepackage{amsmath,stmaryrd}
\usepackage{amsthm}
\usepackage[flushleft]{threeparttable}
\usepackage{array}
\usepackage{booktabs}
\newcommand{\PreserveBackslash}[1]{\let\temp=\\#1\let\\=\temp}
\newcolumntype{C}[1]{>{\PreserveBackslash\centering}p{#1}}
\newcolumntype{R}[1]{>{\PreserveBackslash\raggedleft}p{#1}}
\newcolumntype{L}[1]{>{\PreserveBackslash\raggedright}p{#1}}

\newcommand{\mkv}{-\!\!\!\!\minuso\!\!\!\!-}
\usepackage{bm}
\usepackage{graphicx,subfigure}
\usepackage{algorithm}
\usepackage{cite}
\usepackage{amsmath}
\usepackage{amssymb}
\usepackage{psfrag}
\usepackage{algpseudocode}
\usepackage{empheq}
\usepackage{latexsym, amsmath, subfigure, color, amsfonts, amssymb,graphicx}
\usepackage{wrapfig} 
\newtheorem{Theorem}{Theorem}
\newtheorem{Lemma}{Lemma}
\newtheorem{Definition}{Definition}

\IEEEoverridecommandlockouts
\usepackage{tabularx}

\usepackage[T1]{fontenc}
\usepackage{etoolbox}

\makeatletter
\def\hlinewd#1{%
\noalign{\ifnum0=`}\fi\hrule \@height #1 %
\futurelet\reserved@a\@xhline}
\patchcmd{\maketitle}{\@fnsymbol}{\@alph}{}{}  
\makeatother

\title{The Multi-layer Information Bottleneck Problem}

\author{
    \IEEEauthorblockN{Qianqian Yang\IEEEauthorrefmark{1}, Pablo Piantanida\IEEEauthorrefmark{2}, Deniz G\"und\"uz\IEEEauthorrefmark{1}}
    \IEEEauthorblockA{\IEEEauthorrefmark{1}Electrical and Electronic Engineering Department, Imperial College London, London SW7 2AZ, U.K.
    \\\{q.yang14, d.gunduz\}@imperial.ac.uk}
    \IEEEauthorblockA{\IEEEauthorrefmark{2} Laboratoire des Signaux et Syst\'emes, CentraleSup\'elec-CNRS-Universit\'e Paris-Sud, Gif-sur-Yvette, France
    \\pablo.piantanida@centralesupelec.fr}
}

\date{}

\begin{document}

\maketitle

\begin{abstract}
The muti-layer information\makeatletter{\renewcommand*{\@makefnmark}{}
\footnotetext{This work has been funded in part by the European Research Council (ERC) under Starting Grant BEACON (agreement 677854).}\makeatother} bottleneck (IB) problem, where information is propagated (or successively refined) from layer to layer, is considered. Based on information forwarded by the preceding layer, each stage of the network is required to preserve a certain level of relevance with regards to a specific hidden variable, quantified by the mutual information. The hidden variables and the source can be arbitrarily correlated. The optimal trade-off between rates of relevance and compression (or complexity) is obtained through a single-letter characterization, referred to as the rate-relevance region. Conditions of successive refinabilty are given. Binary source with BSC hidden variables and binary source with BSC/BEC mixed hidden variables are both proved to be successively refinable. We further extend our result to Guassian models.  A counterexample of successive refinability is also provided.
\end{abstract}

\section{Introduction}
A fundamental problem in statistical learning is to extract the relevant essence of data from high-dimensional, noisy, salient sources. In supervised learning (e.g., speaker identification in speech recognition), a set of properties or statistical relationships is pre-specified as relevant information of interest (e.g., name, age or gender of the speaker) targeted to be learned from data; while in unsupervised learning, clusters or low-dimensional representations play the same role. This can be connected to the lossy source compression problem in information theory, where an original source is compressed subject to specifically defined distortion (or loss) with regards to specified relevant information. 

A remarkable step towards understanding the information relevance problem using fundamental information theoretical concepts was made by Tishby \emph{et al.}~\cite{bottleneck1999} with the introduction of the ``information bottleneck '' (IB) method. The relevant information in an observable variable $X$ is defined as the information $X$ can provide about another hidden variable $Y$. The IB framework characterizes the trade-off between the information rates (or complexity) of the reproduction signal $\hat{X}$, and the amount of mutual information it provides about $Y$. The IB method has been found useful in a wide variety of learning applications, e.g., word clustering~\cite{slonim2000document}, image clustering~\cite{goldberger2002unsupervised}, etc. In particular, interesting connections have been recently made between deep learning~\cite{LeCun2015Deep} and the successively refined IB method~\cite{7133169}. 

Despite of the success of the IB method in the machine learning domain, less efforts have been invested in studying it from an information theoretical view. Gilad-Bachrach \emph{et al.}~\cite{gilad2003information} characterize the optimal trade-off between the rates of information and relevance, and provide a single-letter region. As a matter of fact, the conventional IB problem follows as a special instance of the conventional \textit{noisy lossy source coding problem}~\cite{1057738}. Extension of this information-theoretic framework address the collaborative IB problem by Vera \emph{et al.} \cite{DBLP:journals/corr/VeraVP16}, and the distributed biclustering problem by Pichler \emph{et al.}~\cite{7541466}.  Further connections to the problem of joint testing and lossy reconstruction has been recently studied by Katz \emph{et al.}~\cite{7888517}. Also in the information theoretic context, the IB problem is closely related to the pattern classification problem studied in \cite{willems_biometric, tuncel_PR, tuncel_gunduz_PR}; which provides another operational meaning to IB.

In this work, we introduce and investigate the multi-layer IB problem with non-identical hidden variables at each layer. This scenario is highly motivated by deep neural networks (DNN) and the recent work in~\cite{7133169}. Along the propagation of a DNN, each layer compresses its input, which is the output of the preceding layer, to a lower dimensional output, which is forwarded to the next layer. Another scenario may be the hierarchical, multi-layer network, in which information is propagated from higher layers to lower layers sequentially. Users in different layers may be interested in different properties of the original source. The main result of this paper is the full characterization of the rate-relevance region of the multi-layer IB problem. Conditions are provided for successive refinability in the sense of the existence of codes that asymptotically achieve the rate-relevance function, simultaneously at all the layers. Binary source with BSC hidden variables and binary source with mixed BSC$\backslash$BEC hidden variables\footnote{BSC hidden variables are obtained by passing the source through a binary symmetric channel, whereas BEC hidden variables are obtained through a binary erasure channel.} are both proved to successively refinable. The successive refinability is also shown for Guassian sources. We further present a counterexample for which successive refinability no longer holds. It is worth mentioning that the successive refinability of the IB problem is also investigated in \cite{tian2008successive}, with identical hidden variables.  

The rest of the paper is organized as follows. Section II
provides the definitions and presents the main result, the achievability and converse proofs of which are provided in the Appendices. The definition and conditions of successive refinability are shown in Section III. Examples are presented in Section IV. Finally, we conclude the paper in Section V.

\section{Problem formulation}

Let $(X^n, Y_1^n, \dots, Y_L^n)$ be a sequence of $n$ i.i.d. copies of discrete random variables $(X, Y_1, \dots, Y_L)$ taking values in finite alphabets $\mathcal{X}, \mathcal{Y}_1, \dots, \mathcal{Y}_L$, jointly distributed according to $p(x, y_1, \dots, y_L)$, where $X$ is an \textit{observable variable}, while $Y_1, \dots, Y_L$ are \textit{hidden variables} arbitrarily correlated with $X$. 

An $(n, R_1,\dots, R_L)$ \textit{code} for the $L$-layer IB problem, as illustrated in Fig.~1, consists of $L$ encoding functions $f_1, \dots, f_L$, defined as $f_l: \,\mathcal{Z}_{l-1} \rightarrow \mathcal{Z}_l$, $l=1, \dots, L,$ where we set $\mathcal{Z}_0\triangleq \mathcal{X}^n$, and $\mathcal{Z}_l\triangleq [1:2^{nR_l}]$, $l=1, \dots, L$. That is, $R_l$ is the rate of the $l$-th layer encoding function $f_l$, $l=1, \dots, L$, and we assume $R_1\geq R_2\geq \cdots \geq R_L$.

\begin{Definition}(Achievability) For some $R_1\geq \cdots \geq R_L $ and non-negative $\mu_1, \dots, \mu_L$ values, $(R_1,\dots, R_L, \mu_1, \dots, \mu_L)$ is said to be achievable if, for every $\epsilon>0$, there exists an $(n, R_1, \dots, R_L)$-code s.t.
\begin{equation}
 \frac{1}{n}I(Y_l^n; Z_l) \geq \mu_l-\epsilon,~~~~~~~~l=1, \dots, L,  
\end{equation}
for sufficiently large $n$, where $Z_l=f_l(Z_{l-1})$, for $l=1, \dots, L$, and $Z_0 \triangleq X^n$. 
\end{Definition}
The value of $\mu_l$ imposes a lower bound on $I(Y_l^n; Z_l)$, i.e., the relevance with respect to the hidden variable $Y_l$ after $l$-layer encoding of the observable sequence $X^n$. Our goal is to characterize the \textit{rate-relevance region}, $\mathcal{R}$, which is the set of all achievable tuples $(R_1,\dots, R_L, \mu_1, \dots, \mu_L)$. 

\begin{figure}\label{fig:system}
\centering
\includegraphics[width=1\linewidth]{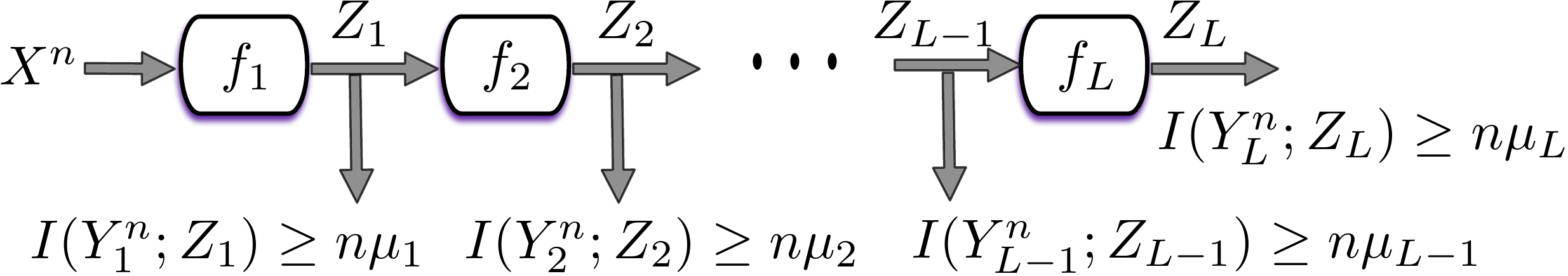}
\caption{Illustration of the multi-layer IB problem.}
\end{figure}

\begin{Theorem}
The rate-relevance region, $\mathcal{R}$, is characterized by the closure of the set of all tuples $(R_1,\dots, R_L, \mu_1, \dots, \mu_L)$ that satisfy
\begin{equation}\label{theorem_1}
R_l\geq I(X;U_l, \dots,  U_{L}),~~~~~l=1, \dots, L,
\end{equation}
for some probability $p(x)p(u_1, \dots, u_L|x)$ s.t.
\begin{equation}\label{theorem_2}
\mu_l\leq I(Y_l; U_l, \dots, U_L),~~~~~l=1, \dots, L.
\end{equation}
\begin{proof}
A proof  is provided in the Appendices. 
\end{proof}
\end{Theorem}  

\section{Successive Refinability of Multi-Layer IB}
The \textit{rate-relevance function} for a single-layer setting with relevance constraint $\mu$ regarding the hidden variable $Y$ is denoted  by $R_{X\rightarrow Y}(\mu)$, and characterized in \cite{gilad2003information} as:
\begin{equation}
R_{X\rightarrow Y}(\mu)=\min\limits_{\substack{ p(U|X):~ U \mkv X \mkv Y,\\I(Y; U)\geq \mu}} I(X;U). \end{equation}
\begin{Definition}
Source $X$ is said to be successively refinable for the $L$-layer IB problem with regards to correlated relevant hidden variables $Y_1, \dots, Y_L$ with relevance constraints $\mu_1, \dots, \mu_2$, respectively, if
\begin{equation}
(R_{X \rightarrow Y_1}(\mu_1),\dots, R_{X \rightarrow Y_L}(\mu_L), \mu_1, \dots, \mu_L) \in \mathcal{R}.  
\end{equation}
\end{Definition}
\begin{Theorem}\label{theo:SR}
Source $X$ is successively refinable for the $L$-layer IB problem with relevance constraints $\mu_1$, \dots, $\mu_L$ with regards to hidden variables $Y_1, \dots, Y_L$, iff there exist random variables $U_1, \dots, U_L$, satisfying  $U_L \mkv  \cdots  \mkv  U_1 \mkv  X \mkv  (Y_1, \dots, Y_L)$, such that the following conditions hold simultaneously for $l=1, \dots, L$:
\begin{enumerate}
\item $R_{X \rightarrow Y_l}(\mu_l)=I(X;U_l)$,
\item $\mu_l\leq I(Y_l; U_l)$.
\end{enumerate}
\begin{proof} Theorem~\ref{theo:SR} follows directly from Definition 2 and Theorem 1.  
\end{proof}
\end{Theorem}

\section{Examples}
\subsection{Binary Source with Symmetric Hidden Variables}\label{section:bi}
We consider $\mathcal{X}=\mathcal{Y}_l=\{0, 1\}$, $l=1, ..., L$. The observable variable $X$ has a Bernoulli distribution  $\frac{1}{2}$ (denoted as Bern $(\frac{1}{2})$), and the hidden variables are obtained by passing the source through independent BSCs, i.e., $Y_l=X\oplus N_l$, where $N_l\sim \mathrm{Bern}(p_l)$, $0\leq p_l\leq \frac{1}{2}$, is independent of $X$, and $\oplus$ denotes modulo-2 addition. 

We first derive the rate-relevance function $R_{X \rightarrow Y_l}(\mu_l)$. Denote by $U_l$ any random variable for which $I(Y_l; U_l)\geq \mu_l$, $U_l\mkv X\mkv Y_l$. We have the following inequality:
\begin{subequations}\label{binary0}
\begin{align}
 \mu_l &\leq H(Y_l)-H(
 X\oplus N_l|U_l)\\
&\leq 1- H_b(p_l*H_b^{-1}(H(X|U_l)))\label{binary0:1}\\
&=1-H_b(p_l*H_b^{-1}(1-I(X;U_l)))\label{binary0:2}
\end{align}
\end{subequations}
where operation $*$ is defined as $a*b=a(1-b)+b(1-a)$, $H_b(\cdot)$ is the binary entropy function, defined as $H_b(p)=p\log 1/p+(1-p)\log \frac{1}{1-p}$, and $H^{-1}_b(\cdot)$ is the inverse of the binary entropy function $H_b(p)$ with $p \in [0, 0.5]$. \eqref{binary0:1} follows from Mrs. Gerber's Lemma and the fact that $H(Y_l)=1$. From \eqref{binary0}, we obtain $I(X;U_l)\geq 1- H_b\left(\frac{H_b^{-1}(1-\mu_l)-p_l}{1-2p_l}\right)$. Thus, we have $R_{X \rightarrow Y_l}{(\mu_l)}\geq 1- H_b\left(\frac{H_b^{-1}(1-\mu_l)-p_l}{1-2p_l}\right)$. Note that by letting $U^*_l=X\oplus M_l$, where $M_l$ is independent of $X$ and $M_l\sim \mathrm{Bern}\left(\frac{H_b^{-1}(1-\mu_l)-p_l}{1-2p_l}\right)$, we have $I(Y_l; U^*_l)= \mu_l$ and $I(X;U^*_l)= 1- H_b\left(\frac{H_b^{-1}(1-\mu_l)-p_l}{1-2p_l}\right)$. We can conclude that $R_{X \rightarrow Y_l}{(\mu_l)}= 1- H_b(\frac{H_b^{-1}(1-\mu_l)-p_l}{1-2p_l})$ and $U^*_l$ given above is a rate-relevance function achieving auxiliary random variable. 
\begin{Lemma}
Binary sources as described above are always successively refinable for the $L$-layer IB problem if $R_{X \rightarrow Y_1}{(\mu_1)}\geq \cdots \geq R_{X \rightarrow Y_L}{(\mu_L)}$ and $\mu_l \leq 1- H_b(p_l), \mathrm{for}~~l=1, ..., L$.
\begin{proof}
Since $R_{X \rightarrow Y_1}{(\mu_1)}\geq \cdots \geq R_{X \rightarrow Y_L}{(\mu_L)}$, we can find binary variables $M_1, ..., M_L$, independent of each other and X, such that $M_1 \oplus\cdots \oplus M_l\sim \mathrm{Bern}(H_b^{-1}(1-R_{X \rightarrow Y_l}{(\mu_l)}))$ for $l=1, ..., L$. By choosing auxiliary random variables: $U_l=X\oplus M_1 \oplus\cdots \oplus M_l$, we have $I(X; U_l)=R_{X \rightarrow Y_l}{(\mu_l)}$ and $I(Y; U_l)=\mu_l$, for $l=1, ..., L$, and $U_L\mkv\cdots\mkv U_1\mkv X\mkv (Y_1, ..., Y_L)$. Together with Theorem 2, this conclude the proof of Lemma 1.  
\end{proof}
\end{Lemma}

\subsection{Binary Source with Mixed Hidden Variables}

Here we consider a two-layer IB problem, i.e., $L=2$. The joint distribution of $(X, Y_1, Y_2)$ is illustrated in Fig.~2, where $X$ is a binary random variable of distribution Bernoulli $\frac{1}{2}$ as in the previous example, but $Y_1$ is the output of a BEC with erasure probability $\epsilon$ ($\epsilon \in [0, 1/2]$) when $X$ is the input, and $Y_2$ is the output of a (BSC) with crossover probability $p$, $ p \in [0, 1/2]$. A similar example can be found in \cite{villard2013secure} where the optimality of proposed coding scheme not always holds for their setting. We first derive the rate-relevance function $R_{X \rightarrow Y_1}(\mu_1)$. Denote by $U_1$ any random variable such that $I(Y_1; U_1)\geq \mu_1$, $U_1\mkv X\mkv Y_1$. We have the following inequality:
\begin{figure}\label{fig:bec_bsc}
\centering
\includegraphics[width=0.8\linewidth]{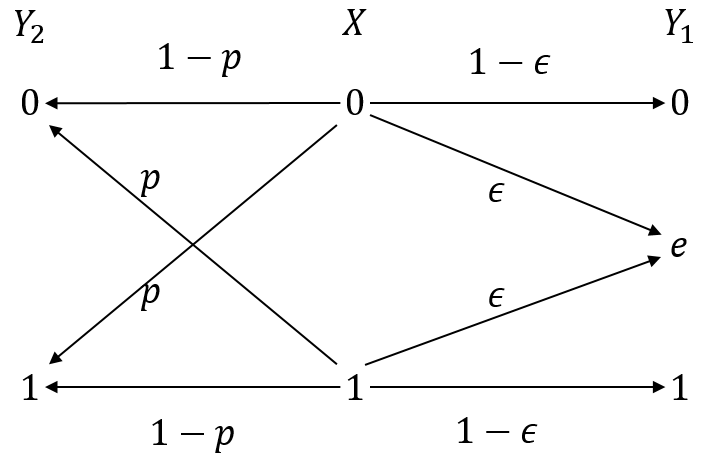}
\caption{Binary source with mixed hidden variables.}
\end{figure}
\begin{subequations}\label{binarybec0}
\begin{align}
&H(
 Y_1| U_1)\nonumber\\
&=H(
 Y_1|U_1)+H(
 X|U_1, Y_1)-H(
 X|U_1, Y_1)\label{binarybec0:1}\\
&=H(
 Y_1, X|U_1)-H(
 X|U_1, Y_1)\label{binarybec0:2}\\
&=H(X|U_1)+H(
 Y_1|X, U_1)-H(
 X|U_1, Y_1)\label{binarybec0:3}\\
&=1-I(X;U_1)+H(
 Y_1|X)\nonumber\\
&~~-\Big(p(Y_1\neq e)H(
 X|U_1, X)\nonumber\\
&~~~~+p(Y_1=e)H(
 X|U_1, Y_1=e)\Big) \label{binarybec0:4}\\
&=1-I(X; U_1)+H_b(\epsilon)-\epsilon H(
 X|U_1) \label{binarybec0:5}
 \\
&=1+(\epsilon-1)I(X;U_1)-\epsilon+H_b(\epsilon).\label{binarybec0:6}
\end{align}
\end{subequations}
Since $\mu_1 \leq I(Y_1, U_1)=H(Y_1)-H(Y_1|U_1)$, where $H(Y_1)=H_e(\epsilon)$, $H_e(\epsilon)$ is defined as $H_e(\epsilon)=\epsilon\log 1/\epsilon +(1-\epsilon)\log \frac{2}{1-\epsilon}$, it follows that $R_{X\rightarrow Y_1}(\mu_1)\geq \frac{\mu_1+H_b(\epsilon)-H_b(\epsilon)+1}{1-\epsilon}$, which can be achieved by setting $U^*_1=X\oplus M_1$, where $M_1\sim \mathrm{Bern}(R_{X\rightarrow Y_1}(\mu_1))$ is independent of $X$. We have $R_{X \rightarrow Y_2}(\mu_2)= 1- H_b\left(\frac{H_b^{-1}(1-\mu_2)-p}{1-2p}\right)$ from Section. \ref{section:bi}, which can be achieved by setting $U^*_2=X\oplus M_2$, where $M_2\sim \mathrm{Bern}(R_{X \rightarrow Y_2}(\mu_2))$ is independent of $X$.

\begin{Lemma}\label{lemma:bec}
Binary source $X$ with mixed BEC/BSC hidden variables as described above is always successively refinable for the $L$-layer IB problem if $R_{X \rightarrow Y_1}{(\mu_1)}\geq \cdots \geq R_{X \rightarrow Y_L}{(\mu_L)}$ and $\mu_l \leq I(X; Y_l)$.
\begin{proof}
The proof follows the same arguments as in the proof of Lemma 1.
\end{proof}
\end{Lemma}

Note that successive refinability is still achievable in this example despite the mixed hidden variables contrast to our expectation, since an auxiliary $U$ in the form $X\oplus M$ achieves the rate-relevance function despite BEC hidden variable.

\subsection{Jointly Gaussian Source and Hidden Variables}

It is not difficult to verify that the above achievability results are still valid for the Gaussian sources by employing a quantization procedure over the sources and appropriate test channels\cite{Gamalnetworkinformation}.

Let $X$ and $Y_l$, $l=1, ..., L$, be jointly Gaussian zero-mean random variables, such that $Y_l=X+N_l$, where $X\sim \mathcal{N}(0, \sigma^2_x)$ and $N_l\sim \mathcal{N}(0, \sigma^2_{N_l})$, $N_l\perp X$. As in the previous examples, we first derive a lower bound on the rate-relevance function $R_{X \rightarrow Y_l}(\mu_l)$. Denote by $U_l$ any random variable such that $I(Y_l; U_l)\geq \mu_l$, $U_l\mkv X\mkv Y_l$. We have the following sequence of inequalities: 
\begin{subequations}\label{gaussian0}
\begin{align}
&\mu_l \leq I(Y_l; U_l)\\ 
&=h(Y_l)-h(
 X+N_l|U_l)\\
&\leq \frac{1}{2}\log\Big(2\pi e (\sigma_{x}^2+\sigma_{N_l}^2)\Big)\nonumber\\  &~~~-\frac{1}{2}\log\Big(2\pi e\sigma_{N_l}^2+\exp{(2h(X|U_l))}\Big)\\
&= \frac{1}{2}\log\Big(2\pi e (\sigma_{x}^2+\sigma_{N_l}^2)\Big)\nonumber\\  &~~~-\frac{1}{2}\log\Big(2\pi e\sigma_{N_l}^2+\exp{(2h(X)-2I(X;U_l))}\Big)\label{gaussian01}\\
&= \frac{1}{2}\log\Big(2\pi e (\sigma_{x}^2+\sigma_{N_l}^2)\Big)\nonumber\\  &~~~-\frac{1}{2}\log\Big(2\pi e\sigma_{N_l}^2+2\pi e\sigma_{x}^2\exp{(-2I(X;U_l)}\Big),
\end{align}
\end{subequations}
where \eqref{gaussian01} follows from the conditional Entropy Power Inequality (EPI) (Section 2.2 in  \cite{Gamalnetworkinformation}). We can also obtain an outer bound on $R_{X\rightarrow Y_l}(\mu_l)$:
\begin{equation}\label{raterelevancefunction}
R_{X \rightarrow Y_l}(\mu_l)\geq \frac{1}{2}\log \frac {2^{2\mu_l} \sigma_x^2}{\sigma_x^2+\sigma_{N_l}^2-2^{2\mu_l}\sigma_{N_l}^2},
\end{equation}
by setting  $U^*_l=X+P_l$, $P_l\sim \mathcal{N}(0, \sigma^2_{P_l})$, $l=1, ..., L$, where $\sigma^2_{P_l}$ is given by:
\begin{align}
\sigma^2_{P_l}=\frac{2\pi e \sigma_x^2-2^{2R_{X \rightarrow Y_l}(\mu_l)}}{2\pi e(2^{2R_{X \rightarrow Y_l}(\mu_l)}-1)}.
\end{align}

\begin{Lemma}
Gaussian sources as described above are always successively refinable for the $L$-layer IB problem if $R_{X \rightarrow Y_1}{(\mu_1)}\geq \cdots \geq R_{X \rightarrow Y_L}{(\mu_L)}$ and $\mu_l \leq \frac{1}{2}\log \frac{2\pi e (\sigma_{x}^2+\sigma_{N_l}^2)}{2\pi e\sigma_{N_l}^2+1}, \mathrm{for}~~l=1, ..., L$.
\end{Lemma}

\subsection{Counterexample on successive refinability}
In this section, we show that the multi-layer IB problem is not always successively refinable. We consider a two-layer IB problem, i.e., $L=2$. Let  $X=(X_1, X_2)$, where $X_1$ and $X_2$ are two independent discrete random variables, and we have $Y_1=X_1$ and $Y_2=X_2$. We first derive the rate-relevance function $R_{X\rightarrow Y_1}(\mu_1)$. Denote by $U_1$ any random variable such that $I(Y_1; U_1)\geq \mu_1$, $U_1-X-Y_1$. We have:
\begin{subequations}\label{counterexample0}
\begin{align}
I(X; U_1)&=I(X_1, X_2; U_1)\\
&\geq I(X_1; U_1)\\
&\geq \mu_1.
\end{align}
\end{subequations}
By setting $U^*_1$ as 
\begin{equation}\label{eq44}
U^*_1=\begin{cases}
X_1, &\mbox{with probability}~~\frac{\mu_1}{H(X_1)},\\
0, &\mbox{with probability}~~1-\frac{\mu_1}{H(X_1)},
\end{cases}
\end{equation}
we have $I(Y_1; U^*_1)= I(X_1; U^*_1)=\mu_1$, and $I(X; U^*_1)=\mu_1$, which achieves the lower bound shown in \eqref{counterexample0}. We can conclude that $R_{X \rightarrow Y_1}(\mu_1)=\mu_1$, and any rate-relevance function achieving random variable $U^*_1$ should satisfy $I(X_2; U^*_1|X_1)=0$, since $I(X; U^*_1)=I(X_1; U^*_1)+I(X_2; U^*_1|X_1)=\mu_1$ and $I(X_1; U^*_1)=\mu_1$. Similarly, we can conclude that $R_{X\rightarrow Y_2}(\mu_2)=\mu_2$, and any rate-relevance function achieving random variable $U^*_2$ should satisfy $I(X_1; U^*_2|X_2)=0$.

\begin{Lemma}\label{lemma:counter}
Source $X$ with hidden variables $Y_1$ and $Y_2$ as described above is not successively refinable for the two-layer IB problem.
\begin{proof}
For any rate-relevance function achieving random variables $U^*_1$ and $U^*_2$, we have 
\begin{subequations}\label{counterexample1}
\begin{align}
&I(U^*_2; X|U^*_1)=I(U^*_2; X_1, X_2|U^*_1)\\
&= I(U^*_2; X_2|U^*_1)+I(U^*_2; X_1|U^*_1, X_2)\\
&\geq I(U^*_2; X_2|U^*_1)\label{counterexample1:1}\\
&=I(U^*_1, U^*_2; X_2)-I(U^*_1; X_2)\\
&=I(U^*_2; X_2)+I(U^*_1; X_2|U^*_2)\label{counterexample1:2}\\
&\geq I(U^*_2; X_2) \geq \mu_2,\label{counterexample1:3} 
\end{align}
\end{subequations}
where \eqref{counterexample1:2} is due to $I(U^*_1; X_2)=0$, which follows from 
\begin{subequations}\label{counterexample11}
\begin{align}
&I(X_1; X_2|U^*_1)\nonumber\\
&=I(X_1, U^*_1; X_2)-I(U^*_1; X_2)\\
&= I(X_2; X_1)+I(X_2; U^*_1|X_1)-I(U^*_1; X_2)\\
&= -I(U^*_1; X_2). 
\end{align}
\end{subequations}
If $\mu_2>0$, $I(U^*_2; X|U^*_1)> 0$, which implies $U^*_2, U^*_1$ and $X$ cannot form a Markov chain for any rate-relevance function achieving random variables $U^*_1$ and $U^*_2$. With Theorem~\ref{theo:SR}, we have proven Lemma~\ref{lemma:counter}.      
\end{proof}
\end{Lemma}

\section{Conclusion}
The multi-layer IB problem with non-identical relevant variables 
was investigated. A single-letter expression of the rate-relevance region was given. The definition and conditions of successive refinability were presented, which was further investigated for the binary sources and Guassian sources. A counterexample of successive refinability was also proposed. 

\bibliographystyle{IEEEtran}
\bibliography{Report}

\appendices

\section{Achivability of Theorem 1}
Consider first the direct part, i.e., every tuple $(R_1,\dots, R_L, \mu_1, \dots, \mu_L)\in \mathcal{R}$ is achievable.  
 
\textbf{Code generation.} Fix a conditional  probability mass function (pmf) $p(u_1, \dots, u_L|x)$ such that $\mu_l$ $\leq$ $I(Y_l; U_l, \dots, U_L)$, for $l$ $=1,\dots,L$. First randomly generate $2^{nR_L}$ sequences $u^n_L(i_L)$, $i_L=[1:2^{nR_L}]$, independent and identically distributed (i.i.d.) according to $p(u_{L})$; then for each $u^n_L(i_L)$ randomly generate $2^{(R_{L-1}-R_L)}$ sequences $u^n_{L-1}(i_L, i_{L-1}),$ $i_{L-1}=[1:2^{n(R_{L-1}-R_L)}]$, conditionally i.i.d. according to $p(u_{L-1}|u_{L})$; and continue in the same manner, for each $u^n_{L-j+1}(i_{L-j+1})$ randomly generate $2^{(R_{L-j}-R_{L-j+1})}$ sequences $u^n_{L-j}(i_L, \dots, i_{L-j})$, $i_{L-j}=[1:2^{n(R_{L-j}-R_{L-j+1})}]$, conditionally i.i.d. according to $p(u_{L-j}|u_{L-j+1}, \dots, u_L)$, for $j=[2: L]$.

\textbf{Encoding and Decoding} After observing $x^n$, the first encoder finds an index tuple $(i_1, \dots, i_L)$ such that $(x^n, u^n_{1}(i_L, \dots, i_1), u^n_{2}(i_L, \dots, i_{2}), \dots, u^n_{L}(i_L))$ is in the set $T^n_{\epsilon}(X, U_1, \dots, U_L)$, which is the set of $\epsilon$ jointly typical $n$ vectors of random variables $X, U_1, \dots, U_L$. If more than one such tuple exist, any one of them is selected. If no such tuple exists,  we call it an error, and set $(i_1, \dots, i_L)=(1, \dots, 1)$. Then the $j$th encoder outputs $(i_{j}, \dots, i_L)$, for $j=1, \dots, L$, and sends to the $j+1$ encoder, if $j<L$, the index tuple $(i_{j}, \dots, i_L)$ at a total rate of $R_j$. Given the index tuple $(i_j, \dots, i_L)$, the $j$th decoder declares $u^n_{1}(i_L, \dots, i_j)$ as its output, for $j=1, \dots, L$.

\textbf{Relevance.} First, we note that if there is no error in the encoding step, i.e., an index tuple $(i_1, \dots, i_L)$ such that $(x^n,$ $u^n_{1}(i_L, \dots, i_1),$ $u^n_{2}(i_L, \dots, i_{2}),$ $\dots,$ $u^n_{L}(i_L))$ $\in T^n_{\epsilon}(X, U_1, \dots, U_2)$ is found, then the relevance condition $\mu_l\leq I(Y; U_l, \dots, U_L)$, $\forall~l=1, \dots, L$, is satisfied by the definition of $T^n_{\epsilon}(X, U_1, \dots, U_L)$ and the Markov lemma. Then we focus on the analysis of the probability of error, i.e., the probability that such an index tuple cannot be found in the encoding step. 

An error occurs if one of the following events happens:
\begin{subequations}
\begin{flalign}
&E_0: x^n \notin T^n_{\epsilon}(X);\\ 
&E_1: x^n \in T^n_{\epsilon}(X), (x^n, u_L^n(i_L)) \notin T^n_{\epsilon}(X, U_L),\nonumber\\ &~~~~~\text{for all}~~ i_L=1, \dots, 2^{R_L};\\
&E_l: (x^n, u_{L-l+2}^n(i_{L-l+2}), \dots, u_L^n(i_L))\nonumber\\
&~~~~~~~\in T^n_{\epsilon}(X, U_{L-l+2}, \dots, U_L),\nonumber\\
&~~~~~(x^n, u_{L-l+1}^n(i_{L-l+1}), \dots, u_L^n(i_L)) \nonumber\\
&~~~~~~~\notin T^n_{\epsilon}(X, U_{L-l+1}, \dots, U_L),\nonumber\\ 
&~~~~~\text{for all}~~i_{L-l+1}=1, \dots, 2^{R_{L-l+1}-R_{L-l+2}};
\end{flalign}
\end{subequations}
for $l=2, \dots, L$. It is clear that $\mathbb{P}(E_0)\rightarrow 0$ as $n \rightarrow \infty$. Based on the properties of typical sequences: 
\begin{subequations}
\begin{flalign}
&\mathbb{P}(E_1)\stackrel{n \rightarrow \infty}{\rightarrow} 0, \text{ if}~~R_{L} \geq I(X; U_{L});\\
&\mathbb{P}(E_l)\stackrel{n \rightarrow \infty}{\rightarrow} 0, \text{ if}~~R_{L-l+1}-R_{L-l+2}\nonumber\\
&~~~~~~~~~~~~\geq I(X; U_{L-l+1}|U_{L-l+2}, \dots, U_{L}),
\end{flalign}
\end{subequations}
for $l=[1:L]$. 

\section{Converse of Theorem 1}
Next, we prove that every achievable tuple $(R_1,\dots, R_L, \mu_1, \dots, \mu_L)$ must belong to $\mathcal{R}$. The system achieving $(R_1,\dots, R_L, \mu_1, \dots, \mu_L)$ is specified by the encoding functions $\{f_1, \dots, f_L\}$, i.e.,
\begin{subequations}
\begin{align}
 &f_1: \mathcal{X}^n \rightarrow \mathcal{Z}_1;\\
 &f_l: \mathcal{Z}_{l-1} \rightarrow \mathcal{Z}_l,~~~l=2, \dots, L,
\end{align}
\end{subequations}
such that 
\begin{subequations}
\begin{align}
 &R_l \geq \frac{1}{n}\log|\mathcal{Z}_{l}|;\\
 &\mu_l \leq \frac{1}{n}I(Y_l^n; Z_{l}), \mbox{for}~l=1, \dots, L.
\end{align}
\end{subequations}
By setting $U_{Li}\triangleq(Z_L, X^{i-1})$, and $U_{li}\triangleq Z_l$, for $i=[1:n]$ and $l=[1:L-1]$, where $X^{i-1}=(X_1, \dots, X_{i-1})$, we have 
\begin{subequations}
\begin{flalign}
nR_l&\geq I(Z_l, \dots, Z_L; X^n)\label{converse1}\\
&=\sum\limits_{i=1}^n I(U_{li}, \dots, U_{Li}; X_i)\label{converse3}\\
&=nI(U_{l}, \dots, U_{L}; X),\label{converse4}
\end{flalign}
\end{subequations}
where \eqref{converse1} is due to the fact that $Z_{l+1}, \dots, Z_L$ are all deterministic function of $Z_l$; \eqref{converse3} follows from the definitions of $U_{li}, \dots, U_{Li}$; and \eqref{converse4} follows by defining $U_l=(U_{lJ}, J), \dots, U_L=(U_{LJ}, J)$, where $J$ is a random variable independent of all other random variables, and uniformly distributed over the set $\{1, \dots, n\}$. We can also write  
\begin{subequations}
\begin{flalign}
n\mu_l&\leq I(Y_l^n; Z_l)\\
&=\sum\limits_{i=1}^n I(Y_{l,i}; Z_l, ..., Z_L, Y^{i-1}_l)\\
&\leq\sum\limits_{i=1}^n I(Y_{l,i}; Z_l, ..., Z_L, Y^{i-1}_l, X^{i-1})\label{converse11}\\
&=\sum\limits_{i=1}^n I(Y_{l,i};Z_l, ..., Z_L, X^{i-1})\\
&~~~+I(Y_{l,i}; Y^{i-1}_l| Z_l, ..., Z_L, X^{i-1})\\
&=\sum\limits_{i=1}^n I(Y_{l,i};U_{li}, ..., U_{Li})\label{eq:iid}\\
&=nI(Y_l; U_l, ..., U_L),
\end{flalign}
\end{subequations}
where \eqref{converse11} is due to the non-negativity of mutual information; and \eqref{eq:iid} follows since $Y_{l,i}\mkv Z_l, ..., Z_L, X^{i-1}\mkv Y^{i-1}_l$ form a Markov chain, which can be proven as follows:
\begin{subequations}
\begin{align}
&I(Y_{l,i}; Y^{i-1}_l|Z_l, ..., Z_L, X^{i-1})\nonumber\\
&=I(Y_{l,i}; Y^{i-1}_l|Z_l, X^{i-1})\\
&=I(Y_{l,i}, Z_l; Y^{i-1}_l |X^{i-1})-I(Z_l; Y^{i-1}_l |X^{i-1})\\
&=I(Y_{l,i}; Y^{i-1}_l |X^{i-1})+I(Z_l; Y^{i-1}_l |X^{i-1}, Y_{l,i})=0,
\end{align}
\end{subequations}
where $I(Z_l; Y^{i-1}_l |X^{i-1})\leq I(Z_{l-1}; Y^{i-1}_l |X^{i-1}) \leq \cdots \leq I(Z_{1}; Y^{i-1}_l |X^{i-1})\leq I(X^n; Y^{i-1}_l |X^{i-1})=0$, and, similarly, $I(Z_l; Y^{i-1}_l |X^{i-1}, Y_{l,i})=0$.

\end{document}